\def\year{2019}\relax
\documentclass[letterpaper]{article} 
\usepackage{aaai}  
\usepackage{times}  
\usepackage{helvet}  
\usepackage{courier}  
\usepackage{url}  
\usepackage{graphicx}  
\usepackage{amssymb,amsmath,amsfonts,bbm,subfigure}
\newcommand{\eg}{e.g.}
\newcommand{\ie}{i.e.}
\newcommand{\etal}{et~al.}
\DeclareMathOperator{\E}{\mathbb{E}}

\begin{document}
%
\title{Adversarial Learning of Semantic Relevance in Text to Image Synthesis}
\author{Miriam Cha, Youngjune L.~Gwon, H.~T.~Kung\\
John A. Paulson School of Engineering and Applied Sciences\\
Harvard University, Cambridge, MA 02138\\}

\maketitle
\begin{abstract}
We describe a new approach that improves the training of generative adversarial nets (GANs) for synthesizing diverse images from a text input. Our approach is based on the conditional version of GANs and expands on previous work leveraging an auxiliary task in the discriminator. Our generated images are not limited to certain classes and do not suffer from mode collapse while semantically matching the text input. A key to our training methods is how to form positive and negative training examples with respect to the class label of a given image. Instead of selecting random training examples, we perform negative sampling based on the semantic distance from a positive example in the class. We evaluate our approach using the Oxford-102 flower dataset, adopting the inception score and multi-scale structural similarity index (MS-SSIM) metrics to assess discriminability and diversity of the generated images. The empirical results indicate greater diversity in the generated images, especially when we gradually select more negative training examples closer to a positive example in the semantic space. 
\end{abstract}

\section{Introduction}
Generative adversarial net (GAN)~\cite{goodfellow2014} has successfully demonstrated the learning of an empirical probability distribution to synthesize a realistic example. Computations for probabilistic inference are often intractable, and GAN provides a viable alternative that uses neural architectures to train a generative model. GAN has been received well and applied to a variety of synthesis tasks \cite{karras2018,Subramanian2017,radford2015unsupervised,reed2016generative}. 

This paper develops a text-to-image synthesis model built on conditional GAN (CGAN) by Mirza \& Osindero \shortcite{mirza2014conditional}. The task of translating a short text description into an image has drawn much interest, attempted in a number of GAN approaches \cite{reed2016generative,tacgan,zhang,stackgan,cha2017,gawwn}. Conditioning both the generator and discriminator on a text description, these approaches are capable of creating realistic images that correspond to the text description given. 

Despite its success, GAN is known to suffer from a \textit{mode collapse} problem in which generated images lack diversity and fall largely into a few trends. One way to mitigate mode collapse is to encourage a bijective mapping between the generated image output and the input latent space \cite{zhu2017,CycleGAN2017}. In Odena \etal~\shortcite{acgan}, auxiliary classifier GAN (AC-GAN) is tasked to recover side information about the generated image such as a class label. The extra task is shown to promote the bijective mapping and discourages different input latent code from generating the same image output. Dash \etal~\shortcite{tacgan} describe text conditioned auxiliary classifier GAN (TAC-GAN). During the training, the auxiliary classifier in TAC-GAN predicts the class of a generated image. The predicted class is compared against the ground-truth class of a training image whose text description is applied as the input.  

\begin{figure}[t]
\centering
\includegraphics[width=0.4\textwidth]{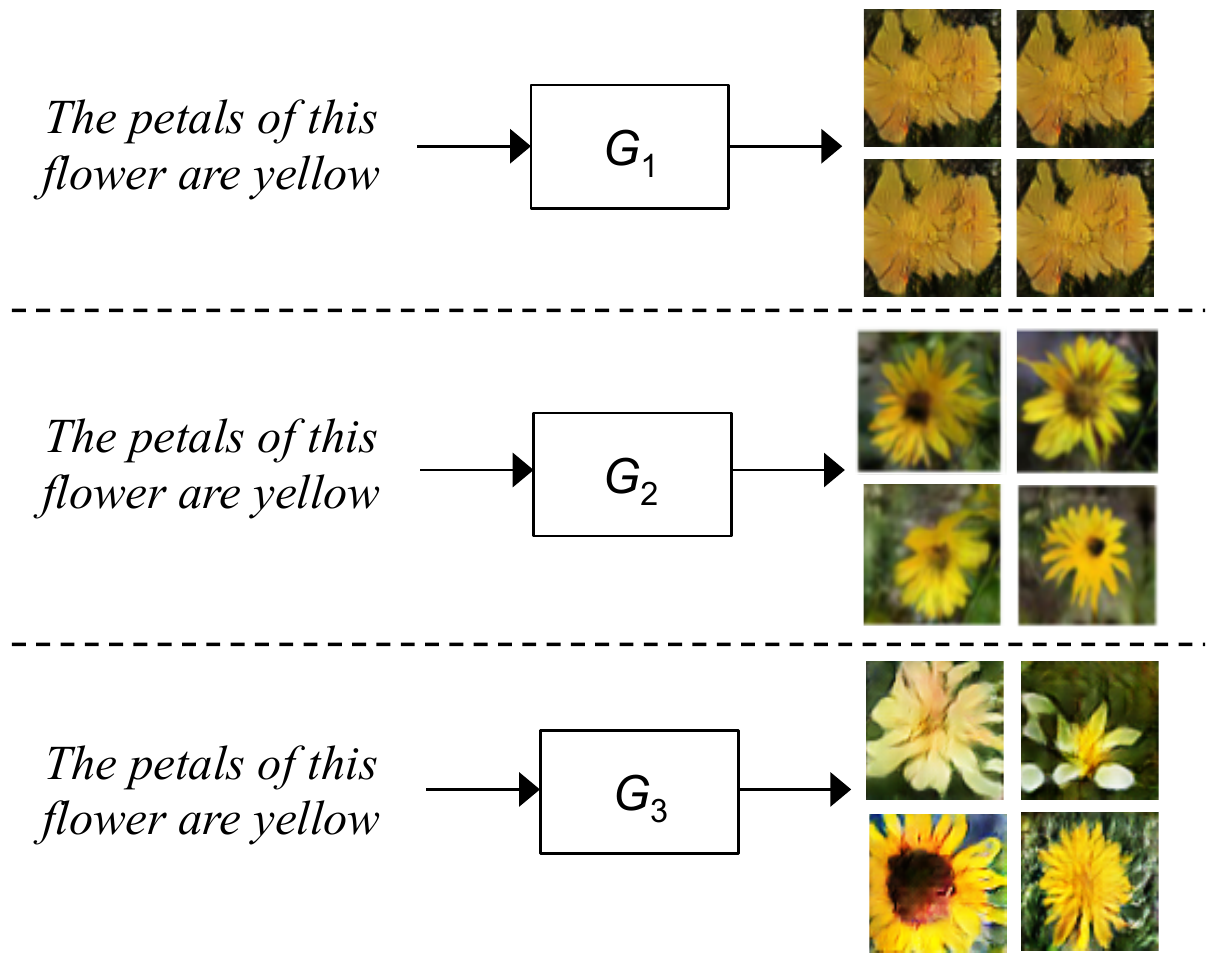}
\caption{An illustration of sample diversity in text-to-image synthesis. Top: poor diversity (generated images are nearly identical suffering from mode collapse); middle: modest diversity (generated images belong to a single class); bottom: good diversity (generated images are not limited to classes)}
\label{fig:motivation} 
\end{figure}

Text is an indiscriminate input choice for CGAN (e.g., compared to class label). The implicit binding of text input to the (predicted) class label output in TAC-GAN is a source of ambiguity because some text can rightfully describe different class of images. Figure~\ref{fig:motivation} illustrates diversity in generated samples. Given a text input, the baseline $G_1$ collapses to synthesizing images of the same trend. When the discriminator is forced to recover a class label of input text, the generator is implicitly bounded to synthesize a single image class as in $G_2$, even if descriptive text is suitable for many different classes of flowers. We are interested in synthesizing images as in $G_3$ that are diverse, realistic, and relevant to the input text regardless of class. 

Instead of class prediction, we modify the discriminator to regress semantic relevance between the two modalities of data, text and image. Similar to the AC-GAN and CGAN text-to-image synthesis, we explore the benefit of an additional supervised task in the discriminator. We train the discriminator with an extra regression task to estimate semantic correctness measure, a fractional value ranging between 0 and 1, with a higher value reflecting more semantic relevance between the image and text. We find training with the extra regression beneficial for the generator diversifying its generated examples, alleviating the mode collapse problem.

To support the learning through semantic correctness, we devise a training method that selects positive and negative examples. Unlike existing approaches that select a random image outside its class as a negative example, we distinguish easy and hard negatives measured in the semantic space of image's text embedding. We validate empirically that our training method significantly improves the diversity of generated images and semantic correctness to the input text.


The rest of this paper is organized as follows. We survey existing approaches of text-to-image synthesis and triplet sampling. Next, we provide a background on GAN and its conditional variants and compare their architectures to our method. We then describe our approach in detail. Using the Oxford-102 flower dataset, our quantitative evaluation compares the discriminability and diversity performance of our method against the state-of-the-art methods. 

\section{Related Work}
CGAN is fundamental to many approaches for text-to-image synthesis. Conditioning gives a means to control the generative process that the original GAN lacks. Reed \etal~\shortcite{reed2016generative} were the first to propose the learning of both generator and discriminator conditioned on text input. They took text description of an image as side information and embedded onto the semantic word vector space for the use in the GAN training. With both generator and discriminator nets conditioned on text embedding, image examples corresponding the description of the text input could be produced. Zhang \etal~\shortcite{stackgan} and Zhang \etal~\shortcite{zhang} improved the quality of generated images by increasing resolution with a two-stage or hierarchically nested CGAN. Other approaches improved on CGAN augment text with synthesized captions \cite{i2t2i} or construct object bounding boxes prior to image generation \cite{gawwn,hong2018}. 

Previous approaches have focused on improving the quality and interpretability of generated images trained on large datasets such as Caltech-UCSD Bird (CUB) \cite{wah2011caltech} and MS-COCO \cite{lin2014microsoft}). Differentiated from the previous work, our primary interest is to remedy the mode collapse problem occurred on the flower dataset \cite{nilsback2008automated} as observed in Reed \etal~\shortcite{reed2016generative}. Notably, Dash \etal~\shortcite{tacgan} make the use of auxiliary classification task to mitigate mode collapse on flower dataset. However, as the same text can be used to describe images of different classes, we suspect that feeding scores on class prediction to generator can potentially bound the generator to produce images of limited number of classes. To solve such problem, we develop a new architecture that uses semantic relevance estimation instead of classification and the training method for increasing the effective usage of limited training examples. 

Unlike previous approaches that form training triplets by randomly selecting negative images, our method selects a negative image based on its semantic distance to its reference text. The idea of selecting negatives for text-image data based on some distance metric is not new, as it has been explored for image-text matching tasks \cite{wang2016}. We gradually decrease semantic distance between reference text and its negative image. The idea of progressively increasing semantic difficulty is related to curriculum learning \cite{curriculum} that introduces gradually more complex concepts instead of randomly presenting training data. Curriculum learning has been successfully applied to GAN in several domains. Subramanian \etal~\shortcite{Subramanian2017} and Press \etal~\shortcite{press2017} use curriculum learning for text generation from gradually increasing the length of character sequences in text as the training progresses. Karras \etal~\shortcite{karras2018} apply curriculum learning on image generation by increasing the image resolution. We are the first to apply curriculum learning based on semantic difficulty for text-to-image synthesis. 
\section{Background}
This section reviews the GAN extensions that condition on or train an auxiliary supervised task with the side information. For text-to-image synthesis, we focus on methods by Reed \etal~\shortcite{reed2016generative} and Dash \etal~\shortcite{tacgan} as they are the closet schemes to ours. We describe our approach in contrast to these extensions. 

\subsection{Conditional GAN (CGAN)}
Goodfellow~\etal~\shortcite{goodfellow2014} suggest a possibility of conditional generative models. Mirza \& Osindero \shortcite{mirza2014conditional} propose CGAN that makes the use of side information at both the generator $G$ and discriminator $D$. A mathematical optimization for CGAN is given by 
{\small \begin{align} \label{eq:cgan} 
\min_G\max_D \E_{\mathbf{x} \sim p_{\text{data}}}[\log D(\mathbf{x}|\mathbf{y})] + \E_{\mathbf{z} \sim p_\mathbf{z}}[\log(1-D(G(\mathbf{z}|\mathbf{y})))]
\end{align}}where $\mathbf{x}$ is a data example, and the side information $\mathbf{y}$ can be a class label or data from another modality. Figure~\ref{fig:gans} (a) shows the structure of CGAN when a class label is used as the side information. GAN data generation requires to sample a random noise input $\mathbf{z}$ from the prior distribution $p_\mathbf{z}$. This approach for multimodal CGAN is convenient and powerful for modalities that are typically observed together (\eg, audio-video, image-text annotation). For practical consideration, $G$ takes in $\mathbf{z}|\mathbf{y}$ as a joint representation that (typically) concatenates $\mathbf{z}$ to $\mathbf{y}$ into a single vector. Similarly, joint representation can be formed for $D$. 

\subsection{Matching-aware Manifold-interpolated GAN (GAN-INT-CLS)}
\begin{figure}[t]
\centering
\includegraphics[width=0.4\textwidth]{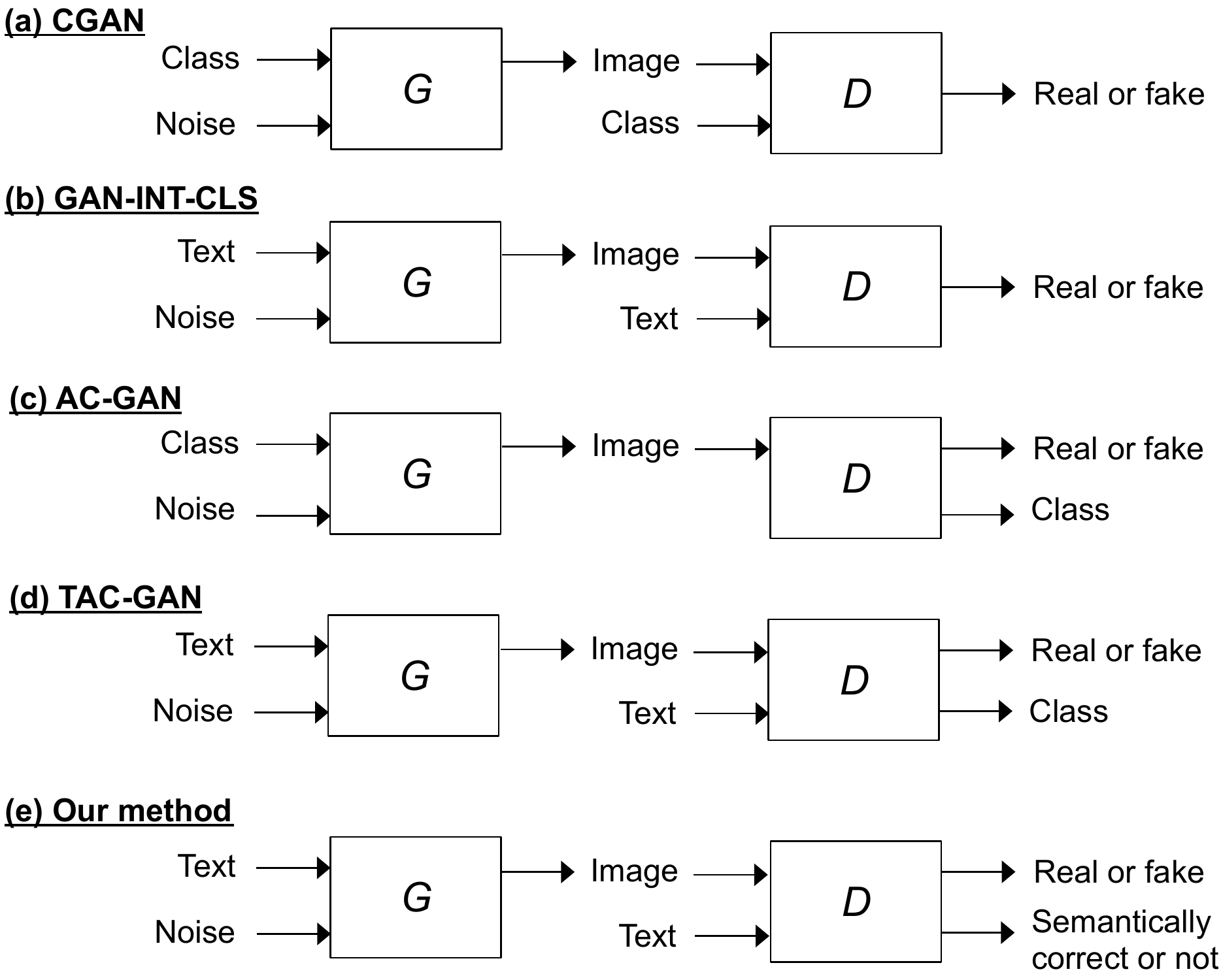}
\caption{Architectural comparison of (a) CGAN \cite{mirza2014conditional}, (b) matching-aware manifold-interpolated GAN \cite{reed2016generative}, (c) auxiliary classifier GAN \cite{acgan}, (d) text conditioned auxiliary classifier GAN \cite{tacgan}, and (e) our method.}
\label{fig:gans} 
\end{figure}

Reed \etal~\shortcite{reed2016generative} propose GAN-INT-CLS for automatic synthesis of realistic images from text input. As in Figure~\ref{fig:gans} (b), GAN-INT-CLS can be viewed as a multimodal CGAN trained on text-annotated images. Text features are pre-computed from a character-level recurrent neural net and fed to both $G$ and $D$ as side information for an image. In $G$, the text feature vector is concatenated with a noise $\mathbf{z}$ and propagated through stages of fractional-strided convolution processing. In $D$, the generated or real image is processed through layers of strided convolution before concatenated to the text feature for computing the final discriminator score. 

\begin{figure*}[t!]
\centering
\includegraphics[width=0.9\textwidth]{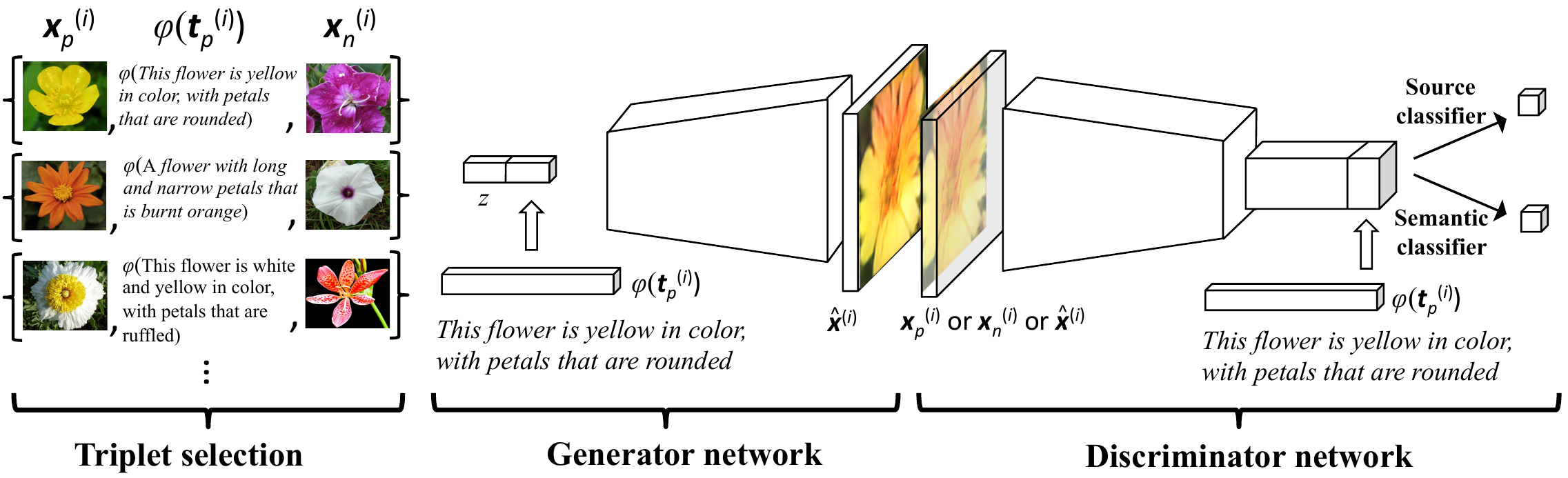}
\caption{Overall pipeline of Text-SeGAN. From training data, we first form mini-batches of $N$ triplet examples $\{\mathbf{x}_p^{(i)}$, $\varphi(\mathbf{t}_p^{(i)})$, $\mathbf{x}_n^{(i)}\}_{i=1}^N$. The formed triplets are used to train the generator and discriminator networks.} 
\label{fig:arc} 
\end{figure*}

\subsection{Auxiliary Classifier GAN (AC-GAN)}
AC-GAN \cite{acgan} uses an extra discriminative task at $D$. For example, the AC-GAN $D$ could perform class label prediction for a generated example from $G$. That is, instead of supplying the class label side information as an input to $D$, AC-GAN trains an additional label classifier at $D$ for input samples. The structure of AC-GAN is illustrated in Figure~\ref{fig:gans} (c). The log-likelihood functions of the AC-GAN $D$ consist of
\begin{equation}
\label{eq:acgan1}
\small
L_s = \E[\log p(s=\textrm{``real''}| \mathbf{x})] + \E[\log p(s=\textrm{``fake''} | \hat{\mathbf{x}})]
\end{equation}
\begin{equation}
\label{eq:acgan2}
\small
L_c = \E[\log p(C=c | \mathbf{x})] + \E[\log p(C=c | \hat{\mathbf{x}})]
\end{equation}
Here, $p(s=\textrm{``real''} | \mathbf{x})$ and $p(s=\textrm{``fake''} | \hat{\mathbf{x}})$ are the source ($s$) probability distributions given a real input $\mathbf{x}$ or generated input $\hat{\mathbf{x}}$. Similarly, $p(C=c | \mathbf{x})$ and $p(C=c |  \hat{\mathbf{x}})$ are the class probability distributions over the labels. During the training, $D$ maximizes $L_s + L_c$ while $G$ maximizes $L_c - L_s$. 

\subsection{Text Conditioned Auxiliary Classifier GAN (TAC-GAN)}
Simply put, TAC-GAN \cite{tacgan} combines AC-GAN and GAN-INT-CLS. Figure~\ref{fig:gans} (d) shows the structure of TAC-GAN.  As in GAN-INT-CLS, TAC-GAN performs image synthesis on the text embedding $\varphi(\mathbf{t})$ with a sampled noise input. Following AC-GAN, the TAC-GAN $D$ performs source and label classifications of the input image. The log-likelihood functions of the source classification and the label classification are given by  
{\small 
\begin{align}
\label{eq:tacgan1}
L_s = \E[\log p&(s=\textrm{``real''} | \mathbf{x}, \varphi(\mathbf{t}))] \nonumber \\
        &+ \E[\log p(s=\textrm{``fake''} | \hat{\mathbf{x}},\varphi(\mathbf{t}))]
\end{align}
\begin{align}
\label{eq:tacgan2}
L_c = \E[\log p(C = c | \mathbf{x},\varphi(\mathbf{t}))] 
        + \E[\log p(C=c | \hat{\mathbf{x}},\varphi(\mathbf{t}))]
\end{align}
}The goal for $D$ is to maximize $L_s + L_c$, and $L_c - L_s$ for $G$. 

The label prediction in $D$ aims to correctly recover the class label of its input. If an input is real, its ground-truth label is available in the training examples. If the input is fake (\ie, generated), its class label is set with the same label as the class of the image associated with the text input to $G$ used to generate the fake image. Thus, $G$ will be penalized if its generated image does not correspond to the class of the image associated with its text input. 

\subsection{Comparison}
All architectures in Figure~\ref{fig:gans} take in the side information input. While CGAN and GAN-INT-CLS discriminators are single-tasked, AC-GAN, TAC-GAN, and our approach have an extra discriminative task. Noticeably, our approach does not rely on class prediction of generated images. This is because there could be a potential one-to-many mapping problem where one text description broadly covers multiple classes of images. As a result, it may cause adverse effect on the generator training. Instead, we wish to weigh in whether or not the input (text description) explains the generated image correctly as shown in Figure~\ref{fig:gans} (e). In the next section, we describe our approach in detail.

\section{Our Approach}

\subsection{Overview}
We propose Text-conditioned Semantic Classifier GAN (Text-SeGAN), a variant of the TAC-GAN architecture and its training strategies. Figure~\ref{fig:arc} illustrates an overview of our approach. Unlike TAC-GAN, there is no class label prediction in the discriminator network. Using a text-image pair from the training dataset, we form a triplet $\{\mathbf{x}_p$, $\varphi(\mathbf{t}_p)$, $\mathbf{x}_n\}$. We denote $\mathbf{x}_p$ a positive image that corresponds to the description of encoded text $\varphi(\mathbf{t}_p)$. On the contrary, a negative image $\mathbf{x}_n$ is selected not to correspond to $\varphi(\mathbf{t}_p)$. Instead of randomly sampling a negative image, we introduce various training strategies that select a negative image based on a semantic distance to its encoded text. Our algorithm for selecting negative images will be explained in detail. We train the generator and discriminator networks using the triplets. By conditioning on the encoded positive text $\varphi(\mathbf{t}_p)$, our generator synthesizes an image $\hat{\mathbf{x}}$ (fake). Taking $\mathbf{x}_p$, $\mathbf{x}_n$, or $\hat{\mathbf{x}}$ as an input conditioned on $\varphi(\mathbf{t}_p)$, the discriminator predicts a source of the input image and semantic relevance to $\varphi(\mathbf{t}_p)$. 

\subsection{Training Objectives}
We train the generator and the discriminator by mini-batch stochastic gradient ascent on their objective functions. Similar to AC-GAN and TAC-GAN, the log-likelihood objective $L_s$ is used to evaluate whether or not a (source) image applied to the discriminator is real or fake
{\small 
\begin{align}
\label{eq:disc_loss}
L_s = \E[\log p&(s=\textrm{``real''} | \mathbf{x}, \varphi(\mathbf{t}))] \nonumber \\
        &+ \E[\log p(s=\textrm{``fake''} | \hat{\mathbf{x}},\varphi(\mathbf{t}))]
\end{align}
}

The additional task at our discriminator is to determine how well the applied image matches the text encoding. In other words, the discriminator is tasked to predict semantic relevance between the applied image and text. The log-likelihood objective $L_r$ for semantic relevance matching is
{\small 
\begin{align}
\label{eq:disc_loss}
L_r = \E[\log p&(r=\textrm{``match''} | \mathbf{x},\varphi(\mathbf{t}))] \nonumber \\
        &+ \E[\log p(r=\textrm{``match''} | \hat{\mathbf{x}},\varphi(\mathbf{t}))]
\end{align}
}Regardless of the applied image source (real or fake), we want to make sure that the image matches the text description. Using the likelihoods $L_s$ and $L_r$, we describe the training objectives for our discriminator and generator. For training the discriminator, we maximize $L_s + L_r$. For training the generator, however, we want to maximize $L_r$ while minimizing $L_s$. For realistic fakes, $\E[\log p(s=\textrm{``fake''} | \hat{\mathbf{x}},\varphi(\mathbf{t}))]$ should be low. Hence, we maximize $L_r - L_s$ for $G$.

\begin{figure*}[t]
\centering
\includegraphics[width=0.98\textwidth]{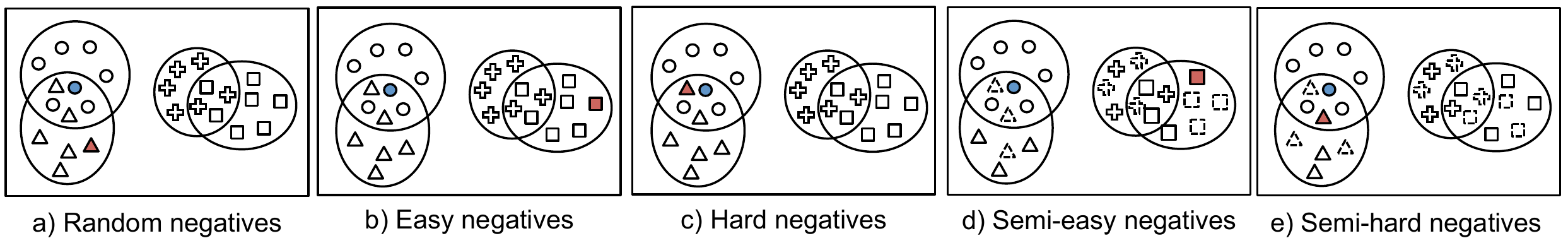}
\caption{Sampling negative images. Different shapes indicate different classes. Images are placed in the semantic space (reduced to 2D for illustrative purposes) where the Euclidean distance is used to indicate separation between points. The selection methods are (a) random negatives, (b) easy negatives, (c) hard negatives, (d) semi-easy negatives, and (e) semi-hard negatives.}
\label{fig:negs} 
\end{figure*}

\subsection{Negative Sampling}
\label{sec:triplet}
Optimizing over all possible triplets is computationally infeasible. To form our triplet $\{\mathbf{x}_p$, $\varphi(\mathbf{t}_p)$, $\mathbf{x}_n\}$, we perform class negative sampling. A positive example is an image with matching text description and a class label drawn from the training dataset. Our class negative sampling looks for an example outside the positive class. Additionally, we use the Euclidean distance between the text embeddings of the reference positive and negative examples.

Recall that our discriminator takes in image and text modalities. The discriminator is applied with the following pairs formed from each triplet. 
\begin{enumerate}
\item Real image with matching text: any reference example as-is from the training dataset is this kind;
\item Real image with non-matching text: any example from negative sampling paired with the reference text description results in this kind;
\item Fake image with matching text: any image generated using the reference text description results in this kind. 
\end{enumerate}

We now discuss methods to select negatives for a given reference (positive) example. Existing text-to-image synthesis approaches \cite{reed2016generative,tacgan,stackgan,cha2017,zhang} select a random image outside the class of the positive example. According to Reed \etal~\shortcite{reed2016generative}, negative images are real images that do not match the reference text description. We note that negative samples could have partial or complete  matching to the reference text but a different label from the reference example. Suppose a reference image $\mathbf{x}^{(i)}$ with a label $l^{(i)}$ has an encoded text description $\varphi(\mathbf{t}^{(i)})$. We define various types of negative images and evaluate their effects on text-to-image synthesis. Figure~\ref{fig:negs} illustrates different negative sampling schemes in the semantic metric space of encoded text. Samples for different classes are denoted by different shapes. Blue circle indicates a reference sample, and one of its negative samples is shown in red. \\
\noindent\textbf{Random negatives.} As used in other approaches \cite{reed2016generative,tacgan,stackgan,cha2017,zhang}, choose any image outside the class of the reference image as a random negative. Given a positive example $\mathbf{x}^{(i)}$, choose
\begin{equation}
\label{eq:rand_neg}
\small
\mathbf{x}^{(j)}~~\text{s.t.}~l^{(i)} \ne l^{(j)},
\end{equation} where $l^{(i)}$ is the class label of $\mathbf{x}^{(i)}$, and $l^{(j)}$ for that of the random negative $\mathbf{x}^{(j)}$. Figure~\ref{fig:negs} (a) illustrates a random negative example (in red triangle). 

\noindent\textbf{Easy negatives.} We use encoded text vectors to measure semantic similarity between images. For easy negative, we find an image that belongs to the outer class of the reference class and has its corresponding text vector farthest from the reference text. We select
\begin{equation}
\label{eq:easy_neg}
\small
\mathbf{x}^{(j)}~~\text{s.t.}~l^{(i)} \ne l^{(j)}~\text{and}~\max_{j} \Arrowvert  \varphi(\mathbf{t}^{(i)}) - \varphi(\mathbf{t}^{(j)})\Arrowvert^2_2~~\forall j
\end{equation}
In Figure~\ref{fig:negs} (b), we note that the red square is the farthest outer class sample from the reference positive image.\\
\noindent\textbf{Hard negatives.} As denoted with red triangle in Figure~\ref{fig:negs} (c), the hard negative corresponds to an image that belongs to the outer class of the reference image and has its text vector closest to the encoded reference text. Thus, we select
\begin{equation}
\label{eq:hard_neg}
\small
\mathbf{x}^{(j)}~~\text{s.t.}~l^{(i)} \ne l^{(j)}~\text{and}~\min_{j} \Arrowvert  \varphi(\mathbf{t}^{(i)}) - \varphi(\mathbf{t}^{(j)})\Arrowvert^2_2~~\forall j\ne i
\end{equation}

\noindent\textbf{Semi-easy negatives.} Selecting easiest negatives in practice leads to a poorly trained discriminator. To mitigate the problem, it helps to select a negative 
\begin{equation}
\label{eq:semi_easy_neg}
\small
\mathbf{x}^{(j)}~~\text{s.t.}~l^{(i)} \ne l^{(j)}~\text{and}~\Arrowvert  \varphi(\mathbf{t}^{(i)}) - \varphi(\mathbf{t}^{(j)})\Arrowvert^2_2 > \alpha
\end{equation}
for some $\alpha$. In practice, we randomly select $M$ samples from outer classes and apply Eq.~(\ref{eq:easy_neg}) among the samples in the outer classes. We call these negative samples semi-easy negatives. In Figure~\ref{fig:negs} (d),  dotted lines indicate samples not included in $M$ outer samples. Among the selected outer samples, the red square represents an easy negative sample. 

\noindent\textbf{Semi-hard negatives.} It is crucial to select hard negatives that can contribute to improving the semantic discriminator. However, selecting only the hardest negatives can result in a collapsed model. Therefore, we apply Eq.~(\ref{eq:hard_neg}) in randomly selected $M$ outer samples. We call these negative examples semi-hard negatives. In Figure~\ref{fig:negs} (e), a semi-hard negative is depicted as the red triangle. 

\noindent\textbf{Easy-to-hard negatives.} In early training, hard negatives that have similar features to the reference example may remove relevant features in representing the positive sample, leading to mode collapse.  As a systematic way to provide negative examples of incremental semantic difficulty, we use curriculum learning \cite{curriculum} by gradually increasing the semantic similarity between the input encoded text and negative image. We use the following method for easy-to-hard negative selection.
\begin{enumerate}
\item Randomly select negative text from $M$ outer samples;
\item Generate a histogram of cosine similarity values between positive and $M$ negative text;
\item Select $100 \beta$-th percentile of the histogram ($0<\beta\leq1$);
\item Increase $\beta$ gradually. 
\end{enumerate}
Low $\beta$ induces the selection of easy negatives, whereas high $\beta$ leads to hard negatives. We sample negative training example from a distribution and continue a sequence of sampling of the distribution which gradually gives more weight $\beta$ to the more semantically difficult negatives, until all examples have equal weight of 1 ($\beta=1$).

\section{Experiments}
We evaluate our models using Oxford-102 flower dataset \cite{nilsback2008automated}. The dataset contains 8,189 flower images from 102 classes, which are names of different flowers. Following Reed~\etal~\shortcite{reed2016generative}, we split the dataset into 82 training-validation and 20 test classes, and resize all images to 64$\times$64$\times$3.

Reed \etal~\shortcite{reed2016learning} provide 10 text descriptions for each image. Each text description consists of a single sentence. For text representation, we use a pretrained character-level ConvNet with a recurrent neural network (char-CNN-RNN) that encodes each sentence into a 1,024-dimensional vector. In our training, we sample $n$ out of 10 sentences and use the average text embedding of the sampled sentences. We use $n=4$ determined empirically. 

\subsection{Implementation Details}
As shown in Figure~\ref{fig:arc}, both the generator and discriminator are implemented as deep convolutional neural nets. We build our GAN models based on the GAN-INT-CLS\footnote{https://github.com/reedscot/icml2016} architecture. We perform dimensionality reduction on the 1,024-dimensional text embedding vectors using a linear projection onto 128 dimensions. The generator input is formed by concatenating the reduced text vector with a 100-dimensional noise sampled from a unit normal distribution. In the discriminator, the text vector is depth-concatenated with the final convolutional feature map. We add an auxiliary classification task at the last layer of the discriminator. The auxiliary task predicts a semantic relevance measure between the input text and image. 

In the model training, we perform mini-batch stochastic gradient ascent with a batch size $N=64$ for 600 epochs. We use the ADAM optimizer \cite{kingma2014adam} with a momentum of 0.5 and a learning rate of 0.0002 as suggested by Radford~\etal~\shortcite{radford2015unsupervised}. We use number of outer samples $M=1000$. We increase $\beta$ from 0.6 to 1 by 0.1 for every 100 epoch. We stay with max $\beta$ once it is reached.   

\subsection{Evaluation Metrics}
We evaluate Text-SeGAN both qualitatively and quantitatively. In our quantitative analysis, we compute the inception score \cite{Salimans2016,Szegedy2015} and the multi-scale structural similarity (MS-SSIM) metric \cite{msssim} for comparative evaluation against other models. The inception score measures whether or not a generated image contains a distinctive class of objects. It also measures how diverse classes of images are produced. The analytical inception score is given by 
\begin{equation}
\label{eq:inception}
\small
IS(G) = \exp(\E_\mathbf{x} D_{KL}(p(y|\mathbf{x})||p(y)))
\end{equation}
where $\mathbf{x}$ is a generated image by $G$, and $y$ indicates the label predicted by the pre-trained inception model \cite{Szegedy2015}. $p(y|\mathbf{x})$ is the conditional class distribution and $p(y)$ is the marginal class distribution. Images that contain distinctive objects will have the conditional class distribution with low entropy. $G$ that outputs diverse class of images will have a marginal class distribution with a high entropy value. Therefore, high KL divergence between the two distributions that leads to high $IS(G)$ is desirable. As suggested by Salimans \etal~\shortcite{Salimans2016}, we evaluate the metric on 30k generated images for each generative model. 

Additionally, we use the MS-SSIM metric to measure interclass diversity of the generated images. In image processing, MS-SSIM is used to indicate similarity of two images in terms of luminance, contrast, and structure. Its use in GAN is primarily for measuring dissimilarity (\ie, diversity) of the generated images \cite{tacgan,acgan,zhang}. A low MS-SSIM value indicates higher diversity or a less likelihood of mode collapsing. Following Odena \etal~\shortcite{acgan}, we sample 400 image pairs randomly within a training class and report their MS-SSIM scores. 


\begin{table}[t]
\centering
\small
\caption{Inception scores of the generated images using random negative sampling}
\label{tab:rand_negs}
\begin{tabular}{r||c}
\hline
\multicolumn{1}{c||}{\textbf{}} & Inception score \\ \hline
GAN-INT-CLS \shortcite{reed2016generative}           & 2.66$\pm$0.03                            \\ \hline
StackGAN \shortcite{stackgan}                & 3.20$\pm$0.01                      \\ \hline
TAC-GAN \shortcite{tacgan}                & 3.45$\pm$0.05                      \\ \hline
HDGAN \shortcite{zhang}                & 3.45$\pm$0.07                      \\ \hline
Text-SeGAN                 & \textbf{3.65$\pm$0.06}                       \\ \hline
\end{tabular}
\end{table}

\begin{table}[t]
\centering
\small
\caption{Inception scores of the generated images from Text-SeGAN using various negative sampling schemes}
\label{tab:all_negs}
\begin{tabular}{r||c}
\hline
                                & Inception score \\ \hline
Hard negatives         & 3.33$\pm$0.03                              \\ \hline
Semi-easy negatives  & 3.69$\pm$0.04                 \\ \hline
Semi-hard negatives  & 3.70$\pm$0.04                        \\ \hline
Easy-to-hard negatives & \textbf{4.03$\pm$0.07}                             \\ \hline
\end{tabular}
\end{table}

\subsection{Quantitative Analysis}
We first evaluate the effect of architectural variations using random negative sampling. We compare Text-SeGAN with GAN-INT-CLS \cite{reed2016generative} and TAC-GAN \cite{tacgan} as they are the closest schemes to ours. Table~\ref{tab:rand_negs} shows the inception scores for GAN-INT-CLS, TAC-GAN, and Text-SeGAN using the random negative sampling scheme. For broader comparison, we also include the results by StackGAN \cite{stackgan} and HDGAN \cite{zhang}. The primary goal of StackGAN and HDGAN is to enhance the resolution of generated images. We achieve a significant improvement over GAN-INT-CLS and competitive results against StackGAN, HDGAN, and TAC-GAN despite the difference in image sizes. Note that our generated images ($64\times64\times3$) are half the size of TAC-GAN ($128\times128\times3$) and a quarter of HDGAN ($256\times256\times3$) and StackGAN ($256\times256\times3$) images. Text-SeGAN improves inception scores of GAN-INT-CLS by 0.99, StackGAN by 0.45, and HDGAN and TAC-GAN by 0.2. It is known that images with higher resolution generally improve discriminability \cite{acgan}. Our improvement in the inception score is significant considering a relatively small size of the generated images. Like HDGAN, it is our future work to increase resolution. 

Next, we evaluate the effects of different triplet selection schemes on the proposed architecture. Table~\ref{tab:all_negs} compares the inception scores for Text-SeGAN using hard, semi-easy, semi-hard, and easy-to-hard negative selection schemes. Easy negative selection turns out choosing negative examples that have a little effect in training of our model. Under easy negative selection, generated images match the text description unreliably, and no visible improvement to the model could be observed. Therefore, we omit reporting its results here. Finding the hardest or the easiest negative of a positive sample evidently results in deterministic pairing. We suspect that a lack of variety in triplets may have led to poor performance in easy and hard negatives. In practice, mislabelled and poorly captured images would dominate the easy and hard negatives. Semi-easy negatives have similar performance to random negatives. Semi-hard negatives further increase the inception score by introducing diverse triplets that contribute more to improving the model. Among various negative sampling schemes, easy-to-hard negative sampling achieves the best performance. This result suggests that training with triplets of gradually more difficult semantics can benefit text-to-image synthesis. 

\begin{figure}[t]
\centering
\includegraphics[width=0.33\textwidth]{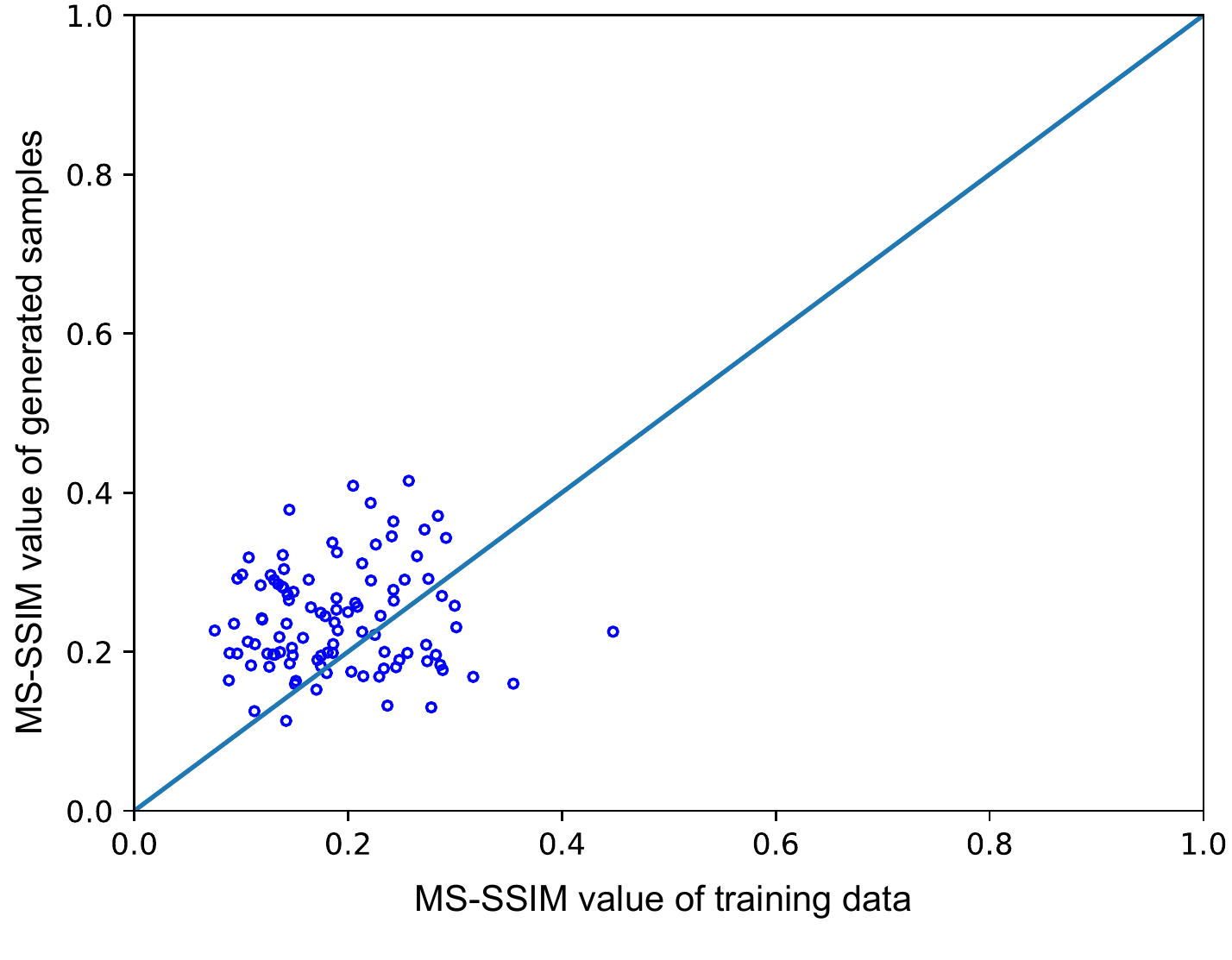}
\caption{Comparison of the class-wise MS-SSIM scores of the samples from the training data and the generated samples of Text-SeGAN using easy-to-hard negative sampling.}
\label{fig:missim} 
\end{figure}

\begin{figure*}[t!]
\centering
\includegraphics[width=0.8\textwidth]{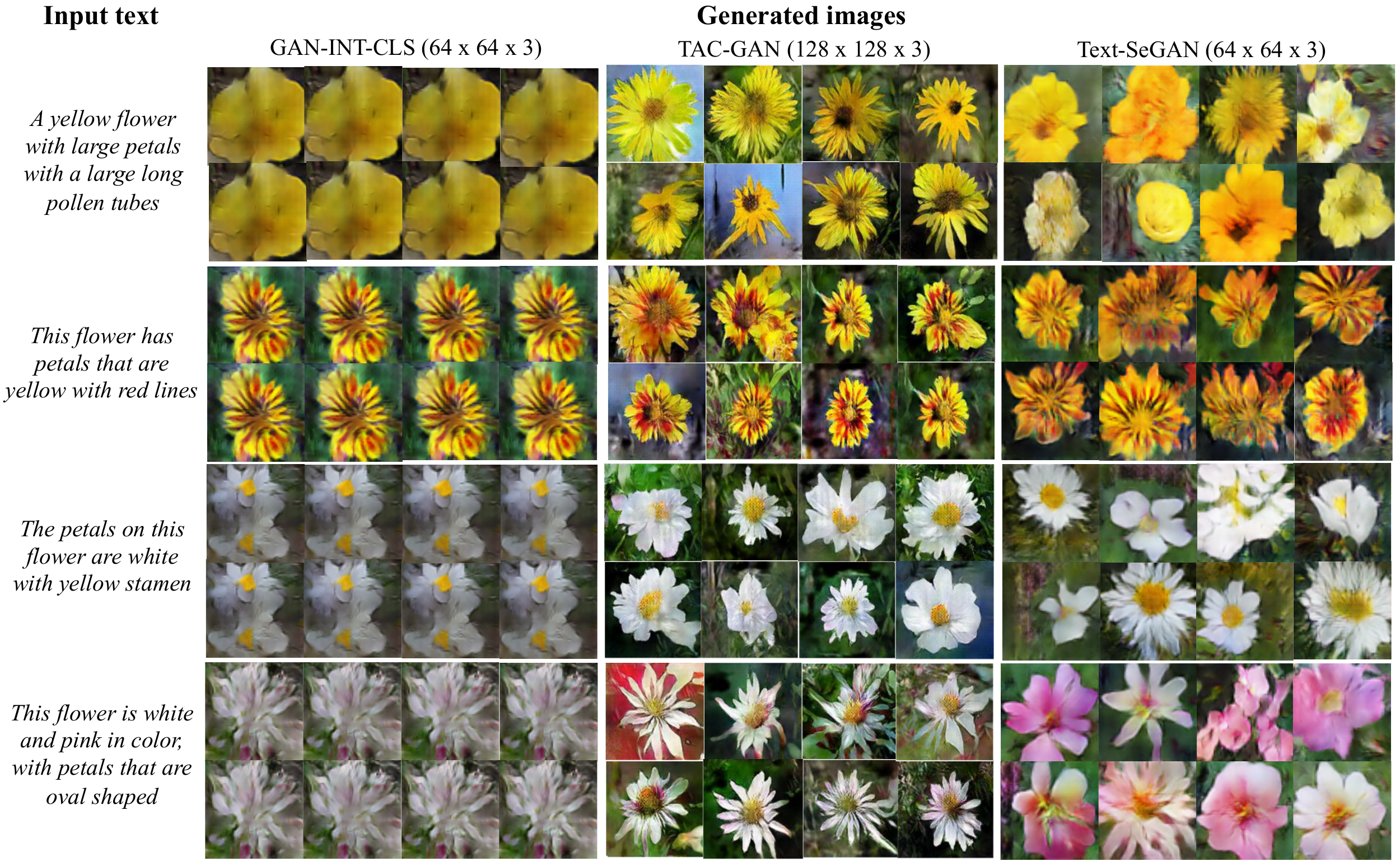}
\caption{Flower images generated by GAN-INT-CLS, TAC-GAN, and Text-SeGAN (with easy-to-hard negatives).}
\label{fig:arc_comp} 
\end{figure*}

As in Dash \etal~\shortcite{tacgan}, we use MS-SSIM to validate that our generated images are as diverse as training data. Figure~\ref{fig:missim} compares the mean MS-SSIM for each class in the training dataset compared to that of the generated images by our best scheme (Text-SeGAN with easy-to-hard negative sampling). Each point represents a class and can be interpreted as how similar the images of the same class are one another in the two datasets. Our model produces as diverse images as the real images in the training dataset. 


\subsection{Qualitative Analysis}

Figure~\ref{fig:arc_comp} shows the generated images from the three text-conditioned models, GAN-INT-CLS ($64\times64\times3$), TAC-GAN ($128\times128\times3$), and Text-SeGAN ($64\times64\times3$) using easy-to-hard negative sampling. At first glance, all images seem reasonable, matching the whole or part of the text descriptions. Despite the semantic relevance and visual realism, we notice that GAN-INT-CLS collapses to nearly an identical image output. In other words, different latent codes are mapped to a few (or singular) output trends. TAC-GAN avoids such collapse, but the generated images tend to belong to a single  class. Adopting an auxiliary class prediction suppresses different latent codes being mapped to the same output, but enforcing the generated image classes to match the class of the input text in TAC-GAN has restricted generating images of diverse classes. Since the goal of text-to-image synthesis is simply generating images that match the input text, we modify the auxiliary classifier to measure semantic relevance instead of reinforcing the class label of the text description attached to training images. In addition, we gradually introduce semantically more difficult triplets rather than in a random order during training. With such modifications, we observe that the generated flowers have more variations in their shapes and colors (in spite of their relative small size of $64\times64\times3$) while matching the text description. For example, given an input text, ``This flower is white and pink in color, with petals that are oval shaped,'' our approach generates diversely shaped flowers with both pink and white petals.

\section{Conclusion and Future Work}
We present a new architecture and training strategies for text-to-image synthesis that can improve diversity of generated images. Discriminator in existing AC-GAN is tasked to predict class label along with source classification (real or fake). Due to one-to-many mapping (the same text can describe images of different classes) in text-to-image synthesis, feeding scores on class prediction to generator can potentially bound the generator to produce images of limited number of classes. In order to mitigate this, we introduce a variant of AC-GAN whose discriminator measures semantic relevance between image and text instead of class prediction. We also provide several strategies of selecting training triplet examples. Instead of randomly presenting the triplets during training, we introduce gradually more complex triplets in their semantics. Experiment results on Oxford-102 flower dataset demonstrate that our model with easy-to-hard negative selection scheme can generate diverse images that are semantically relevant to the input text and significantly improves inception score compared to existing state-of-the-art methods. In future work, we plan to increase resolution of the generated images and further develop methods of training data selection.\\
{\small
\textbf{Acknowledgments.} This work is supported by the MIT Lincoln Laboratory Lincoln Scholars Program and the Air Force Research Laboratory under agreement number FA8750-18-1-0112. This work is also supported in part by a gift from MediaTek USA.
}

\clearpage
\bibliography{paper}
\bibliographystyle{aaai}

\end{document}